\PassOptionsToPackage{colorlinks=true, linkcolor=blue, citecolor=blue, urlcolor=blue}{hyperref}
\documentclass[letterpaper, 10 pt, journal, twoside]{IEEETran}

\IEEEoverridecommandlockouts                              
\makeatletter
\def\endthebibliography{%
	\def\@noitemerr{\@latex@warning{Empty `thebibliography' environment}}%
	\endlist
}
\makeatother

\usepackage{graphics}
\usepackage{epsfig}
\usepackage{siunitx}
\usepackage{verbatim}
\usepackage{mathrsfs}
\usepackage{booktabs}
\usepackage{multirow}
\usepackage{upgreek}
\usepackage{color}
\usepackage[table]{xcolor} 
\usepackage{pifont}
\usepackage{orcidlink}

\usepackage{cuted}
\usepackage{capt-of}

\usepackage{graphics} 
\usepackage{epsfig} 
\usepackage{times} 
\usepackage{amsmath} 
\usepackage{amssymb}  
\usepackage{siunitx}
\usepackage{verbatim}

\usepackage{booktabs}
\usepackage{multirow}
\usepackage{upgreek}
\usepackage{color}
\usepackage{orcidlink}
\usepackage[table]{xcolor}
\usepackage{pifont}
\usepackage{xcolor}

\usepackage[ruled,vlined]{algorithm2e} 

\title{\LARGE \bf
SI-Diff: A Framework for Learning Search and High-Precision Insertion with a Force-Domain Diffusion Policy}

\author{Yibo Liu \href{https://orcid.org/0000-0003-1143-3242}{\includegraphics[width=8pt]{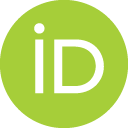}}, Stanko Oparnica, Simon Shewchun-Jakaitis, \\Guoyi Fu, Jie Wang, Jun Yang, Anand Jagannathan, Tony Hong-Yau Lo
 \thanks{Manuscript received: October 10, 2025; Revised: February 5, 2026; Accepted April 25, 2026. This paper was recommended for publication by
Editor Ashis Banerjee upon evaluation of the Associate Editor and Reviewers’
comments. \textit{(Corresponding author: Yibo Liu.)}}%
\thanks{The authors are with Epson Canada, Markham, Ontario L3R 6G3, Canada. (email: buaayorklau@gmail.com)}
\thanks{Simon Shewchun-Jakaiti is with Queen’s University, Kingston, Ontario K7L 3N6, Canada. He made contributions during his internship at Epson Canada.}
\thanks{Project page: https://yorklyb.github.io/SI-Diff/}
\thanks{Digital Object Identifier: see top of this page.}}

\begin{document}
\maketitle
 \markboth{IEEE ROBOTICS AND AUTOMATION LETTERS. PREPRINT VERSION. ACCEPTED
APRIL, 2026}%
{Liu \MakeLowercase{\textit{et al.}}: SI-Diff: A Framework for Learning Search and High-Precision Insertion with a Force-Domain Diffusion Policy}

\begin{abstract}
Contact-rich assembly is fundamental in robotics but poses significant challenges due to uncertainties in relative poses, such as misalignments and small clearances in peg-in-hole tasks. Existing approaches typically address search and high-precision insertion separately, because these tasks involve distinct action patterns.   
However, supporting both tasks within a single model, without switching models or weights, is desirable for intelligent assembly systems.
In this work, we propose SI-Diff, a framework that learns both search and high-precision insertion through a force-domain diffusion policy. To this end, we introduce a new mode-conditioning mechanism that enables the policy to capture distinct action behaviors under a single framework. Moreover, we develop a new search teacher policy that can generate diverse trajectories. By training on successful and efficient demonstrations provided by the teacher policy, the model learns the mapping from tactile and end-effector velocity observations to effective action behaviors. We conduct thorough experiments to show that SI-Diff extends the tolerance to x–y misalignments from 2 mm to 5 mm compared to the state-of-the-art baseline, TacDiffusion \cite{tac}, while also demonstrating strong zero-shot transferability to unseen shapes.
\end{abstract}
\begin{IEEEkeywords}
Force and Tactile Sensing; Assembly;  Force Control; Diffusion Policy.
\end{IEEEkeywords}

\section{INTRODUCTION} \label{intro}

 \IEEEPARstart{C}{ontact-rich} assembly \cite{tac,tree,forge,inoue2017} is crucial in robotics, especially for manufacturing and service robot systems designed to replace human labor.
In real-world scenarios, imprecise placement of objects or parts leads to uncertainties in relative poses, which pose significant challenges for automated assembly \cite{uncertain}.
Although visual servoing \cite{tao} can achieve coarse alignment, its accuracy in cost-constrained industrial environments is typically limited to the millimeter level, which is insufficient for sub-millimeter-clearance peg-in-hole tasks. Thus, force-based methods are required to reliably complete the insertion.
Addressing these challenges requires different strategies tailored to specific types of pose uncertainty.
Taking the peg-in-hole task as an example, when x–y misalignment is the main concern, search strategies \cite{search,Wang_2023} are employed to align the peg and hole. 
The objective of search is to detect the hole boundary and reduce positional error through lateral exploration.
When the peg and hole are generally aligned but the clearance is small, smart insertion strategies \cite{tree,tac,inoue2017} are required to accomplish high-precision (sub-millimeter-level) tasks.
The goal of insertion is to guide the peg to its desired displacement/pose within the hole while resolving unfavorable contact conditions, such as jamming, during the process.
\par
However, most previous research either focuses solely on the search \cite{search, Wang_2023} or the high-precision insertion problem \cite{tac}, or adopts separate methodologies to address them individually \cite{tree}.
Developing a framework to address both tasks is challenging, given that search and high-precision insertion entail distinct action patterns.
In contrast, an experienced human worker, effectively functioning as a single framework, can perform two tasks without delay in switching knowledge domains.
The capability to learn and execute diverse strategies within a single framework is advantageous for real-world industrial applications, as it reduces engineering complexity and facilitates integration compared to separately designed modules.
\par
Recently, TacDiffusion \cite{tac} was proposed to tackle relative pose uncertainties in high-precision peg-in-hole tasks based on a force-domain diffusion policy. 
It adopts a feedforward-force-based impedance control framework, where the feedforward force corresponds to a smart wiggle motion learned from a rule-based teacher policy \cite{tree}, enabling the peg to escape from a stuck state. 
This wiggle motion involves 6-DoF end-effector (EE) actions, suggesting potential for addressing both search and high-precision insertion simultaneously.
However, TacDiffusion assumes the peg and hole are already generally aligned and cannot cope with misalignments beyond 2 mm, indicating limited tolerance to pose variations in practice.
\par
\begin{figure*}[t]
  \centering
  \includegraphics[width=17cm]{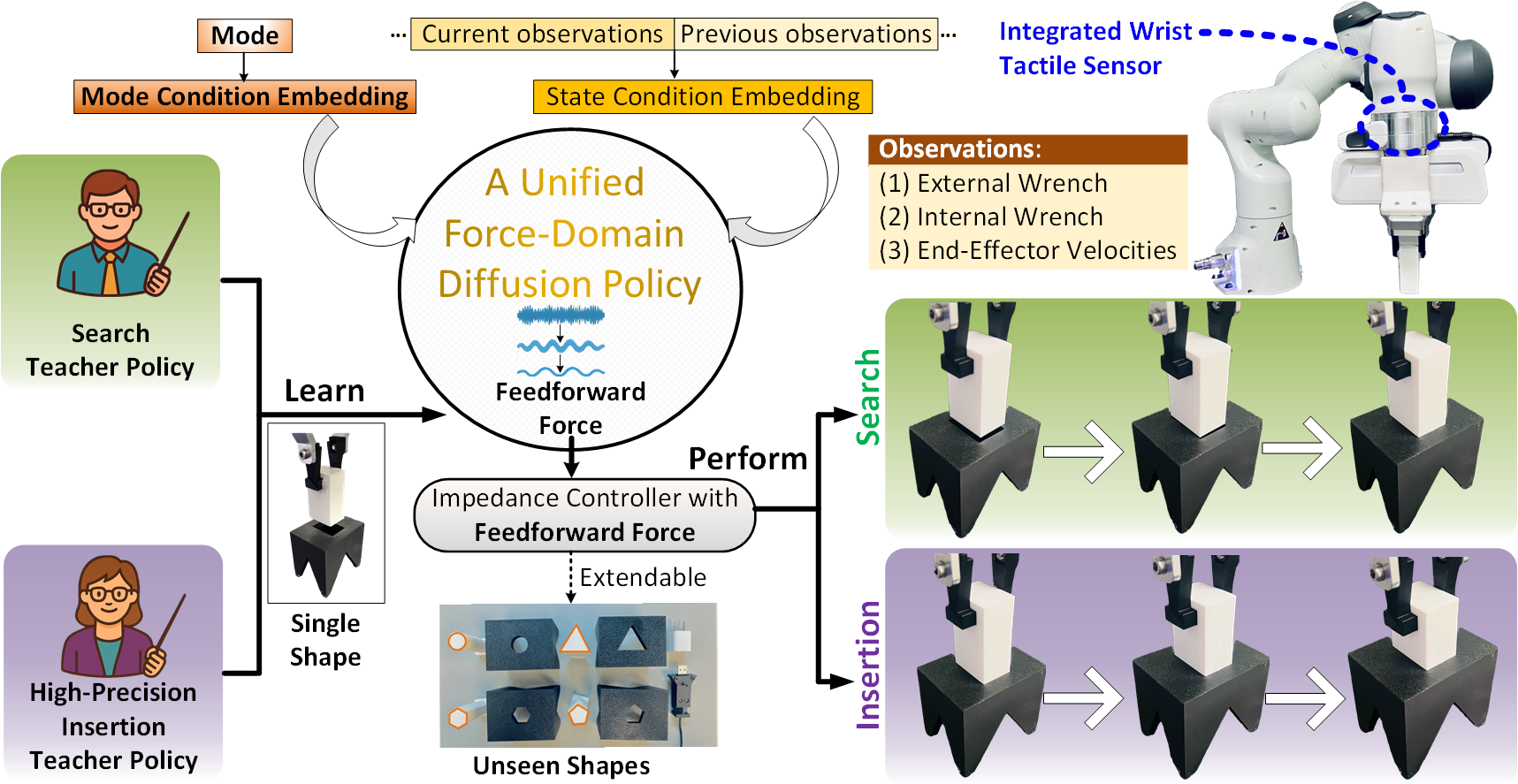}
  \caption{Framework Overview. SI-Diff takes tactile and spatial observations, together with a mode prompt, as input and generates feedforward forces to drive an impedance controller in executing distinct action patterns for search and insertion. This is achieved by simultaneously learning from two teacher policies through the proposed mode conditioning mechanism. Although SI-Diff is trained on a single shape, it demonstrates zero-shot generalization to unseen shapes.}
  \label{overview}
\end{figure*}
In this work, we propose SI-Diff, a novel framework that learns search and high-precision insertion with a force-domain diffusion policy. As shown in Fig. \ref{overview}, our method learns and performs two distinct action behaviors under a single framework.
Compared to TacDiffusion \cite{tac}, which only learns to escape from stuck states, we develop a new rule-based teacher policy that handles non-trivial misalignments and enables efficient training data collection by generating diverse trajectories.
We develop a new mode conditioning mechanism to guide the policy in learning action patterns for search and insertion simultaneously and sufficiently.
We conduct thorough experiments to demonstrate that, compared to the state-of-the-art baseline model TacDiffusion \cite{tac}, our method extends the error tolerance ability from 2 mm to 5 mm. Moreover, SI-Diff only learns from a single shape but shows robust zero-shot transferability to unseen peg shapes.
\par
The \textbf{contributions} of this work are:
\begin{itemize}
\item We develop a novel framework for learning search and high-precision insertion with a force-domain diffusion policy. 
\item {We design a new mode conditioning mechanism, which enables the model to learn two tasks simultaneously while performing distinct action patterns based on the mode prompt.}
\item {We develop an effective search teacher policy that can generate diverse trajectories, enabling efficient training data collection.}
\end{itemize}

\section{Related Work} \label{rel}
\subsection{High-Precision Peg-in-hole Task} 
Although peg-in-hole has been extensively studied, most prior works regard insertion after peg–hole alignment as a simple task \cite{zhang2021,Wang_2023} that can be accomplished by merely applying a force in the insertion direction, which in practice is realized through velocity or position controllers.
For example, in \cite{zhang2021}, the insertion primitive is to simply apply a consistent velocity command along the z-direction. The other example is \cite{Wang_2023}, where the insertion is completed through a position-based controller.
Unfortunately, as introduced in \cite{tac,tree,inoue2017}, when the clearance between the peg and hole is small, relying solely on velocity or position controllers is insufficient, as the tolerance and the robot’s precision are at the same level (\textit{e.g.} sub-millimeter level). Even advanced localization systems based on vision \cite{liaoauto,shuo2,ghostdeblur,shuo} and LiDAR \cite{ifm,ifm2} struggle to reliably achieve this level of accuracy, making contact-based feedback essential.
To address the task of high-precision insertion, \cite{inoue2017,forge} propose learning force-domain control strategies instead of velocity- or position-based control methods. 
However, these studies \cite{inoue2017,forge} focus on specific small-clearance setups with known peg geometries, without demonstrating generalization to unseen tasks \cite{survey}. 
Furthermore, in the absence of a peg model, it must be obtained via reconstruction \cite{unigaussian,lpr} or generative modeling approaches \cite{mvdeepsdf,hippo,yanglearning} prior to deployment.
%
\cite{yan} proposes learning insertion from observations of fingertip tactile sensors and demonstrates good generalization ability, while the use of fingertip tactile sensors in industrial production may face practical considerations due to the cost under frequent physical contact.
Using only a tactile sensor integrated into the robot wrist, 1kHz-BT \cite{tree} introduces a self-adaptive tactile insertion method that employs a contact estimator to detect when the peg is stuck and a human-like wiggling strategy to escape from it, demonstrating decent transferability to unseen shapes.
Since the method is decoupled from the peg shape, it can be extended to unseen shapes. TacDiffusion \cite{tac} is a follow-up to 1kHz-BT \cite{tree} that proposes a force-domain diffusion policy trained exclusively on the successful demonstrations from 1kHz-BT and outperforms it in terms of both success rate and transferability.
Unfortunately, TacDiffusion \cite{tac} assumes that the peg and hole are generally aligned at the beginning and focuses only on completing the insertion thereafter. Hence, it has very limited tolerance to initial misalignment, which hinders its practical application. The aim of this work is to bridge this technological gap.

\subsection{Search Strategies for Peg-in-Hole}
The aim of the search is to eliminate the initial misalignment between the peg and hole prior to insertion \cite{search}. 
In contrast to high-precision insertion \cite{tree,tac}, which primarily involves motion along the insertion direction supplemented by wiggling motions in other directions, the search phase exhibits distinctly different behaviors as it focuses on exploration in the x–y plane.
A simple yet effective solution \cite{search} is to follow a predefined search trajectory. Some studies improve search efficiency using vision sensors \cite{vision1,vision2}, but these typically require carefully positioned cameras to observe the peg and hole. 
Other works have explored force-based search strategies, but some rely on inefficient search patterns \cite{search2}, while others are tailored to specific geometric configurations \cite{active}.
In this work, inspired by 1kHz-BT \cite{tree} and TacDiffusion \cite{tac}, we aim to design a search solution that is decoupled from the peg shape \cite{hippo}. By introducing stochasticity into the search actions and training solely on successful demonstrations, our model acquires diverse action strategies and learns to respond appropriately to varying tactile observations, thereby completing the search task more efficiently.

\section{Methodology} 

\subsection{Overview System Design} \label{secoverview}
Our system learns and executes two distinct behaviors within a single framework, requiring consistent representations during both training and inference. We next describe how this requirement guides our design. The second-order dynamic model of an $\boldsymbol{n}$-Degree-of-Freedom (DoF) torque-controlled robot \cite{tac} is as follows:
\begin{equation}
\mathbf{M}(\boldsymbol{q})\,\ddot{\boldsymbol{q}} 
+ \mathbf{C}(\boldsymbol{q}, \dot{\boldsymbol{q}})\,\dot{\boldsymbol{q}} 
+ \boldsymbol{g}(\boldsymbol{q}) 
= \boldsymbol{\tau}_m + \boldsymbol{\tau}_{\mathrm{ext}}, \label{dynamic}
\end{equation}
where $\boldsymbol{q}\in \mathbb{R}^n$ denotes the joint states, and $\mathbf{M}(\boldsymbol{q}) \in \mathbb{R}^{n\times n}$ denotes the mass matrix. $ \mathbf{C}(\boldsymbol{q}, \dot{\boldsymbol{q}}) \in \mathbb{R}^{n\times n}$ is the Coriolis matrix. $\boldsymbol{g}(\boldsymbol{q}) \in \mathbb{R}^n$ is the gravity vector. $\boldsymbol{\tau}_m \in \mathbb{R}^n$ represents the joint torque, which serves as the control input. $\boldsymbol{\tau}_{\mathrm{ext}} \in \mathbb{R}^n$ represents the external torque applied to the joints.
\par
In this work, we employ a simplified impedance controller with feedforward-force \cite{feed,tac} to regulate the motion of the end-effector (EE):
\begin{equation}
\boldsymbol{\tau}_m = \mathbf{J}(\boldsymbol{q})^\top \big( \boldsymbol{F}_{ff} + \mathbf{K}\boldsymbol{e} + \mathbf{D}\dot{\boldsymbol{e}} \big)
+ \mathbf{C}(\boldsymbol{q}, \dot{\boldsymbol{q}})\dot{\boldsymbol{q}} 
+ \boldsymbol{g}(\boldsymbol{q}), \label{controller}
\end{equation}
where $\mathbf{J}(\boldsymbol{q})^\top \in \mathbb{R}^{n\times 6}$ is the transposed Jacobian matrix that maps wrenches in the 6-DoF Cartesian space of the EE frame to torques in the $n$-DoF joint space. $\boldsymbol{F}_{ff} \in \mathbb{R}^6$ is the feedforward-force \cite{feed}. $\mathbf{K} \in \mathbb{R}^{6\times6}$ and $\mathbf{D} \in \mathbb{R}^{6\times6}$ denote the stiffness and damping matrices, respectively. $\boldsymbol{e} \in \mathbb{R}^6$ represents the error between the current and desired EE poses.
\par
As shown in Eq. (\ref{controller}), EE motion is regulated via $\boldsymbol{F}_{ff}$ or $\boldsymbol{e}$. While search can be achieved by either method, small-clearance insertion requires $\boldsymbol{F}_{ff}$ to overcome jamming \cite{inoue2017,tree,tac}. Since $\boldsymbol{F}_{ff}$ is applicable to both tasks, we adopt it as the shared learnable representation and model output. Considering that the success of two tasks indicates the EE reacts to tactile observations by performing appropriate motions, we adopt the external wrench applied to the EE, $\boldsymbol{\tau}^{e}_{\mathrm{ext}} \in \mathbb{R}^6$; the internal wrench applied to the EE, $\boldsymbol{\tau}^{e}_{\mathrm{in}} \in \mathbb{R}^6$; and the EE velocities in the local EE frame, $\boldsymbol{v}^{e} \in \mathbb{R}^6$, as the shared model input. In particular, $\boldsymbol{\tau}^{e}_{\mathrm{ext}}$
and $\boldsymbol{\tau}^{e}_{\mathrm{in}}$ are defined as:
\begin{equation}
\begin{aligned}
&\boldsymbol{\tau}^{e}_{\mathrm{ext}} = [\boldsymbol{F}_{ext}^\top, \boldsymbol{M}_{ext}^\top]^\top, \quad \boldsymbol{F}_{ext}^\top, \boldsymbol{M}_{ext}^\top \in \mathbb{R}^3, \\
&\boldsymbol{\tau}^{e}_{\mathrm{in}} = [\boldsymbol{F}_{in}^\top, \boldsymbol{M}_{in}^\top]^\top, \quad \boldsymbol{F}_{in}^\top, \boldsymbol{M}_{in}^\top \in \mathbb{R}^3.  \label{extindef}
 \end{aligned}
\end{equation} \par
They are given by:
\begin{equation}
\begin{aligned}
& \boldsymbol{\tau}^{e}_{\mathrm{ext}} = \mathbf{J}(\boldsymbol{q}) \boldsymbol{\tau}_{\mathrm{ext}}, \\
& \boldsymbol{\tau}^{e}_{\mathrm{in}} = \mathbf{J}_{binv}^\top\big(\boldsymbol{\tau}_m - \mathbf{C}(\boldsymbol{q}, \dot{\boldsymbol{q}})\,\dot{\boldsymbol{q}} 
- \boldsymbol{g}(\boldsymbol{q}) \big),  \\
\end{aligned} \label{extin}
\end{equation}
where $\mathbf{J}(\boldsymbol{q}) \in \mathbb{R}^{n\times 6}$ is the Jacobian matrix that maps torques in the $n$-DoF joint space to wrenches in the 6-DoF Cartesian space of the EE frame. $\mathbf{J}_{binv}^\top \in \mathbb{R}^{n\times 6}$ is the Jacobian matrix that relates $n$-DoF joint velocities to 6-DoF EE twist expressed in the EE frame \footnote{If we compare Eq. (\ref{controller}) and Eq. (\ref{extin}), we find that the theoretical value of $\boldsymbol{\tau}^{e}_{\mathrm{in}}$ is $\boldsymbol{F}_{ff} + \mathbf{K}\boldsymbol{e} + \mathbf{D}\dot{\boldsymbol{e}}$. 
Nevertheless, it is preferable to follow Eq. (\ref{extin}) in practice, as it yields the measured $\boldsymbol{\tau}^{e}_{\mathrm{in}}$.}.
\par
In summary, both the search and insertion modes in this work follow the controller defined in Eq. (\ref{controller}), and we rely on learning different $\boldsymbol{F}_{ff}$ profiles to execute different modes.

\begin{figure}[t]
	\centering
	\includegraphics[width=3.3in]{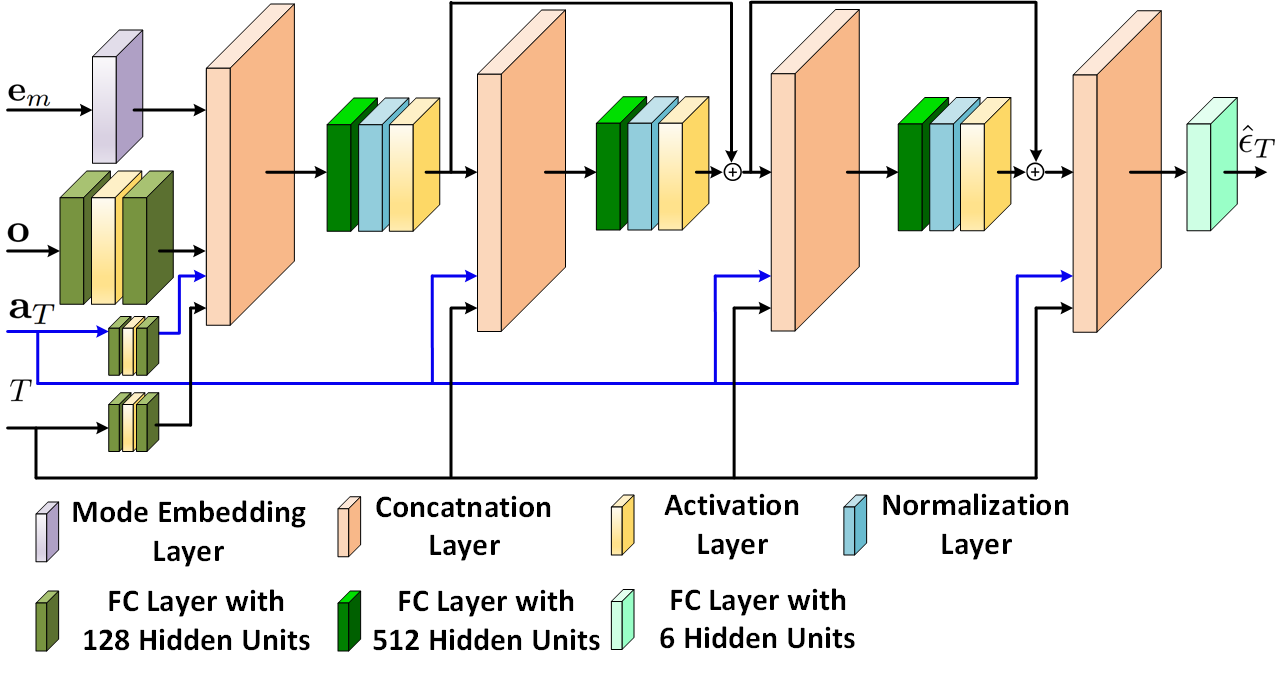}
	\caption{An illustration of the architecture of the proposed force-domain diffusion policy with mode conditioning.
    }
  
	\label{network}
\end{figure}
\subsection{Force-domain Diffusion Policy with Mode Conditioning}
\subsubsection{Force-domain Diffusion Policy}
We employ a Denoising Diffusion Probabilistic Model (DDPM) \cite{DDPM,tac} to learn $\boldsymbol{F}_{ff}$.
The DDPM maps Gaussian noise to the target distribution, which corresponds to the desired $\boldsymbol{F}_{ff}$ for search and high-precision insertion. A DDPM is composed of two stages: the diffusion process and the denoising process.
\par
In the diffusion process, given a clean action sample $\mathbf{a}_{0}$ obtained from training demonstrations, it is transformed into Gaussian noise $\mathbf{a}_{N} \sim \mathcal{N}(\mathbf{0}, \mathbf{I})$ by progressively adding noise  ($\mathbf{a}_{0}\rightarrow \mathbf{a}_{N}$):
\begin{equation}
\begin{aligned}
&\mathbf{a}_T = \sqrt{{\alpha}_T}\,\mathbf{a}_{T-1} + \sqrt{1 - {\alpha}_T}\,\epsilon_{T}, \\
&\alpha_T = 1 - \beta_T,
\end{aligned}
\end{equation}
where the subscript $T = 0,1,\cdots,N$ represents the $T$-th diffusion step. $\epsilon_T \sim \mathcal{N}(\mathbf{0},\mathbf{I})$ is added random noise at $T$-th step. ${\alpha}_{T}$ and $\beta_T$ are predefined parameters for the noise scheduler. Then, in the denoising process, we train a neural network $\hat\epsilon(\cdot)$ to predict the noise added at each diffusion step by minimizing:
\begin{equation}
\mathcal{L}_{\mathrm{DDPM}} = \mathbb{E}\left[\left\| \hat{\epsilon}_{T}(\mathbf{e}_{m}, \mathbf{o}, \mathbf{a}_{T}, T) - \epsilon_{T} \right\|_{2}^{2}\right],
\end{equation}
where $\mathbf{e}_m \in \mathbb{R}^{128}$ is the mode embedding, the details of which will be introduced in Eq. (\ref{eqembed}). $\mathbf{o} \in \mathbb{R}^{36}$ denotes the observations, acting as the condition for denoising. In this work, $\mathbf{o}$ includes the states (6-DoF $\boldsymbol{\tau}^{e}_{\mathrm{ext}}$, 6-DoF $\boldsymbol{\tau}^{e}_{\mathrm{in}}$, and 6-DoF $\boldsymbol{v}^{e}$) of both the current and previous frames. $\hat{\epsilon}_{T}(\cdot)$ and $\epsilon_{T}$ denote the estimated and actual noise at the $T$-th denoising step, respectively. During inference, given Gaussian noise $\mathbf{a}_{N} \sim \mathcal{N}(\mathbf{0}, \mathbf{I})$, the trained model $\hat{\epsilon}(\cdot)$ is leveraged to recover the desired noise-free robot action $\mathbf{a_{0}}$ by iteratively conducting the denoising process ($\mathbf{a}_{N}\rightarrow \mathbf{a}_{0}$):
\begin{equation}
\begin{aligned}
&\mathbf{a}_{T-1} = \frac{1}{\sqrt{\alpha_T}} \left[ \mathbf{a}_T - \frac{1 - \alpha_T}{\sqrt{1 - \bar{\alpha}_T}} \, \hat{\epsilon}_T(\mathbf{e}_m, \mathbf{o}, \mathbf{a}_T, T) \right] + \sigma_T \epsilon_T, \\
& \bar{\alpha}_T = \prod_{i=1}^{T} \alpha_i, \quad \sigma_T = \sqrt{\beta_T}.
\end{aligned}
\end{equation} 
\subsubsection{Mode Conditioning Mechanism} \label{modecon}
We follow the method introduced in TacDiffusion \cite{tac} to inject the conditions of robot states ($\boldsymbol{\tau}^{e}_{\mathrm{ext}}$, $\boldsymbol{\tau}^{e}_{\mathrm{in}}$, and $\boldsymbol{v}^{e}$) into the denoising process. However, TacDiffusion focuses only on learning a single action mode, whereas in this work, our model needs to learn two distinct action behaviors. Thus, it is necessary to design a new condition-injection mechanism to guide the model in learning and performing these actions. 
Inspired by word tokenization \cite{clip} in natural language processing, we consider the modes as a dictionary:
\begin{equation}
\mathcal{D} = \{0,1\},
\end{equation}
where $0$ and $1$ correspond to search and high-precision insertion, respectively. 
Since the mode indices are discrete categories, the model cannot learn efficiently from them directly, 
as it may misinterpret them as having ordinal or numerical magnitude.
Therefore, inspired by the context injection in diffusion-model-based image generation \cite{vqadiff}, we embed the mode indices into high-dimensional learnable vectors:
\begin{equation}
\mathbf{e}_m = \mathbf{E}[m], \quad m \in \mathcal{D}, \label{eqembed}
\end{equation}
where $\mathbf{E} \in \mathbb{R}^{2 \times 128}$ is the learnable embedding matrix 
and $\mathbf{e}_m \in \mathbb{R}^{128}$ is the embedding vector corresponding to index $m$. Inspired by how the \textit{Classification Token} (an additional token for classification) is prepended in BERT \cite{bert} and ViT \cite{vit}, we likewise prepend the learnable mode token to the sequence of robot-state tokens. Fig. \ref{network} shows the architecture of the proposed force-domain diffusion model with mode conditioning.
\begin{figure}[t]
	\centering
	\includegraphics[width=3.3in]{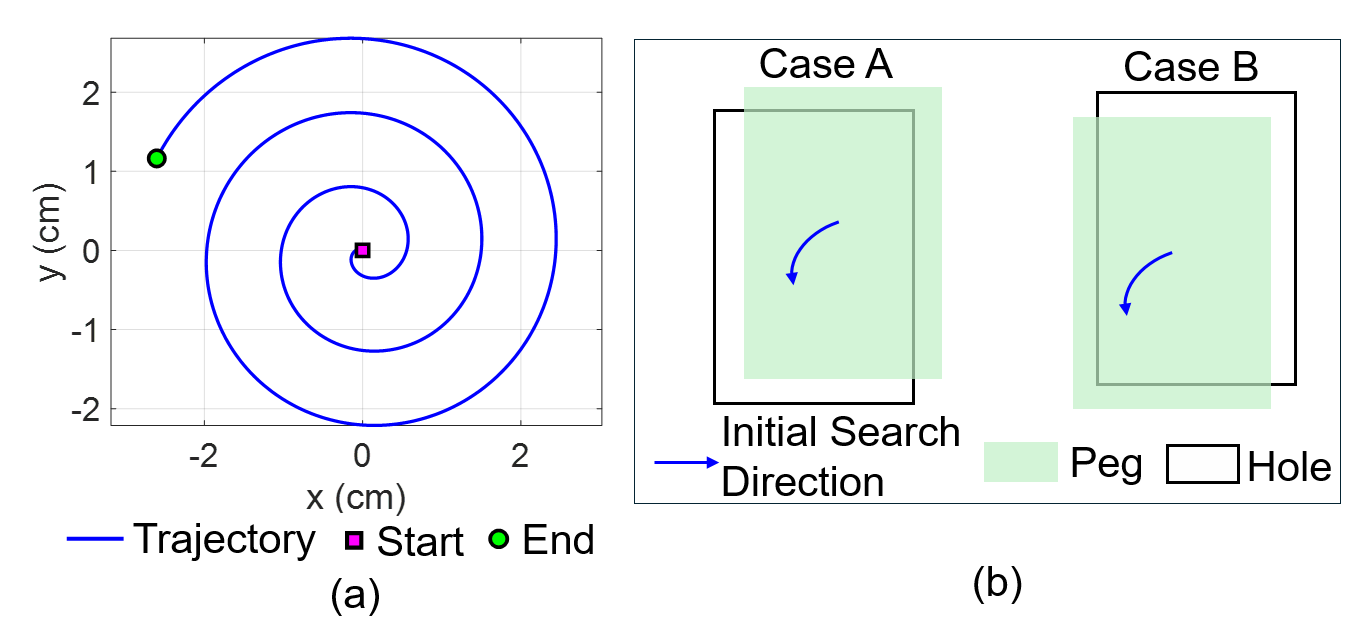}
	\caption{
  (a) An illustration of the trajectory of the vanilla spiral search. (b) The same initial search direction can lead to different efficiency and success outcomes, as it may either guide the peg toward the hole (Case A) or drive it away (Case B), depending on the initial misalignment.
    }
	\label{vanilla}
 
\end{figure}

\subsection{Search Teacher Policy} \label{searchsec}
The objective of the search is to eliminate the initial misalignment in the x–y plane. Following the analysis in Sec. \ref{secoverview}, the teacher policy for search should be driven by  $\boldsymbol{F}_{ff}$. The vanilla spiral search can be achieved by applying the following $\boldsymbol{F}_{ff} = [f_x, f_y, f_z, 0, 0, 0]^\top$ to EE:
\begin{equation}
\left\{
\begin{array}{l}
f_x(\theta) = r(\theta)*\sin(\theta) \\
f_y(\theta) = r(\theta)*\cos(\theta) \\
f_z = c \\
r(\theta) = a + b*\theta \label{eqsearch}
\end{array}
\right.
\end{equation}
where $a$, $b$, and $c$ are predefined constants. $\theta$ increases with time. $r(\theta)$ is the scale for the searching range. As $|r(\theta)|$ increases, the search spans a progressively larger area. 
Although vanilla spiral search is simple, it has clear limitations. As shown in Fig. \ref{vanilla}(a), once $a$, $b$, and $\theta$ are fixed, the trajectory is fully determined. As illustrated in Fig. \ref{vanilla}(b), this may either guide the peg toward the hole (Case A) or away from it (Case B), resulting in inefficiency or failure. Moreover, since Eq. (\ref{eqsearch}) defines actions independently of EE wrenches, it cannot serve as an effective teacher for learning wrench–action mappings (Table \ref{tab2}, first row).

\par
To address the aforementioned problems caused by the fixed searching pattern, we develop a new search policy that introduces randomized parameters together with dedicated design choices, thereby increasing action diversity and improving adaptability. The step-by-step procedure of the proposed search policy is summarized in \textbf{Algorithm \ref{algo1}}. 
\begin{algorithm}[h]
\textbf{Initialization:}\\  \label{algo1}
{\color{gray}\#Select search pattern} \\
$p_s \sim \mathcal{U}\{-1,1\}$ \\
 {\color{gray}\#Select sign for $f_x$ and $f_y$} \\
$d_x \sim \mathcal{U}\{-1,1\}$, $d_y \sim \mathcal{U}\{-1,1\}$ \\
 {\color{gray}\# Set the termination condition} \\
 $z_{max}$ = 3 mm \\
 {\color{gray}\#Initialize parameters for search} \\
$a\sim \mathcal{U}[-0.2,-0.1]$, $b \sim \mathcal{U}[-0.2,-0.1]$\\
$c\sim \mathcal{U}[-10.0,-5.0]$, $\theta \sim \mathcal{U}[-40.0,-60.0]$\\
$\Delta\theta \sim \mathcal{U}[-1.0e^{-2},-5.0e^{-2}]$, $f_z =c$, $\Delta f_z =2.0e^{-3}$

\vspace{3mm}
\textbf{Control Loop:}\\ 
    \While{$ z < z_{max}$}{
    $r = a + b*\theta$ \\
    \If{$p_s$==$\mathrm{1}$}{
    $f_x = d_x*r*\sin(\theta)$, $f_y = d_y*r*\cos(\theta)$
}
\ElseIf{$p_s$==$\mathrm{-1}$}{
   $f_x = d_x*r*\cos(\theta)$, $f_y = d_y*r*\sin(\theta)$
}
Measure external force $f^{ext}_{z}$  \\
\If{$f^{ext}_{z} > -c + \mathrm{0.05}$}{
    $f_z = f_z + \Delta f_z$
}
\ElseIf{$f^{ext}_{z} < -c - \mathrm{0.05}$}{
   $f_z = f_z - \Delta f_z$
} 
$\boldsymbol{F}_{ff} = [f_x, f_y, f_z, 0, 0, 0]^\top$ \\
Apply the controller introduced in Eq. (\ref{controller}) \\
$\theta = \theta + \Delta\theta$ 
    } 

\vspace{3mm}
\textbf{Termination:}\\ 
Maintain the last $\boldsymbol{F}_{ff}$ for 0.25~s. \\
\vspace{3mm}
\textbf{End}. \\
    \caption{Proposed Search Policy} 
    
\end{algorithm} \par

\par
\begin{figure}[h]
	\centering
	\includegraphics[width=3.3in]{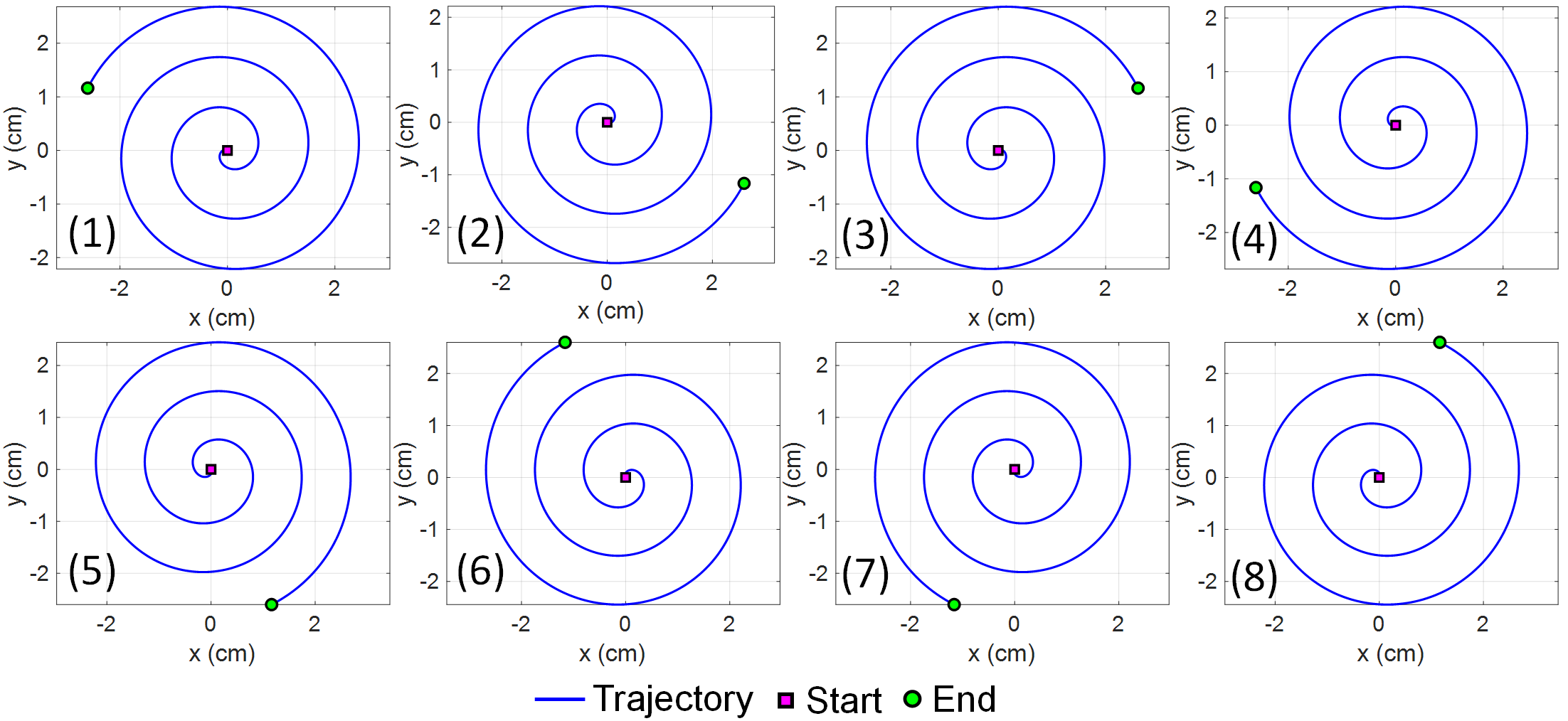}
	\caption{Our search policy can perform eight different trajectories depending on the initialization of random parameters, whose effectiveness varies with different initial misalignments. From left to right and from top to bottom, the subfigures illustrate: (1) $p_s=1$, $d_x=1$, $d_y=1$; (2) $p_s=1$, $d_x=-1$, $d_y=-1$; (3) $p_s=1$, $d_x=-1$, $d_y=1$; (4) $p_s=1$, $d_x=1$, $d_y=-1$; (5) $p_s=-1$, $d_x=1$, $d_y=1$; (6) $p_s=-1$, $d_x=-1$, $d_y=-1$; (7) $p_s=-1$, $d_x=-1$, $d_y=1$; (8) $p_s=-1$, $d_x=1$, $d_y=-1$. These figures are for illustration only, as the actual trajectories also vary with the random initialization of parameters $a$, $b$, $\theta$, and $\Delta \theta$.
    } \label{searchall}
\end{figure}

In Algorithm \ref{algo1}, $z$ denotes the displacement of the EE along its local 
z-axis, measured with respect to the initial position in the EE frame. The notation $v\sim \mathcal{U}(\cdot)$ indicates that 
$v$ is sampled uniformly at random from a continuous interval or from a discrete set.
$f^{ext}_{z}$ is the external force applied to EE along the z-axis of the EE frame. 
As shown in Fig. \ref{searchall}, by designing the initialization of $p_s$, $d_x$, and $d_y$, the proposed policy can perform eight different trajectories. In addition, the random initialization of $a$, $b$, $\theta$, and $\Delta \theta$ introduces diversity in the search behaviors, ranging from gentle to aggressive. For $f_z$, we maintain consistent contact between the peg and table so that the x–y forces/torques are informative of misalignments that the policy can effectively learn from. The random initialization of $c$ covers diverse contact states, reducing out-of-distribution risks during testing. To allow the model to learn when to stop, the last $\boldsymbol{F}_{ff}$ is maintained for 0.25 s after $z$ reaches $z_{max}$. Finally, in Eq. (\ref{controller}), the desired EE pose is fixed to the initial one when the search starts.
\par
The effectiveness of the trajectories generated by our search policy varies with the initial misalignment. Although it is not guaranteed that each generated trajectory is sufficient for a given misalignment, we retain only those samples where the search completes successfully and efficiently. The criterion used to judge whether a demonstration is efficient is whether the search is completed within 2.5 seconds. 
%
%

Consequently, the demonstrations from the teacher policy correspond to appropriate trajectory–misalignment pairs, from which the model can learn to map external and internal wrenches and EE velocities to the search action space. 

\subsection{High-precision Insertion Teacher Policy}
The objective of high-precision insertion in this work is to achieve the desired peg displacement with a small clearance, assuming that the search has aligned the peg and hole. We adopt the self-adaptable tactile insertion method introduced in \cite{tree} as the insertion teacher policy. In particular, the residual force $\boldsymbol{F}_{res}$ is monitored to assess whether the peg is stuck:
\begin{equation}
 \boldsymbol{F}_{res} = \boldsymbol{F}_{in} - \boldsymbol{F}_{ext}
 \label{stuck}
\end{equation}
where $\boldsymbol{F}_{in}$ and $\boldsymbol{F}_{ext}$ are the 3-DoF internal and external forces applied to EE (refer to Eq. (\ref{extindef})). Define the component of  $\boldsymbol{F}_{res}$ in z-direction as $f_{res,z}$. At the beginning of insertion, a reference insertion speed is recorded when $f_{res,z}$ reaches a local maximum, corresponding to a non-stuck state. When the peg’s speed falls below this reference, the peg is regarded as stuck, and the following $\boldsymbol{F}_{ff}$ is applied to the EE to escape the stuck condition:
\begin{equation}
\boldsymbol{F}_{ff,j}(t) = A_j \cdot \sin\!\left( 2\pi f_j t + \varphi_j \right)
\end{equation}
where $A_j$, $f_j$, and $\varphi_j$ denote amplitude, frequency, and phase, respectively. The subscript $j$ represents the direction in the range $x, y, rx, ry, rz$ of the EE frame. Note that in this task, according to the controller in Eq. (\ref{controller}), $\boldsymbol{F}_{ff}$ is only responsible for releasing the peg from being stuck, while the insertion is driven by $\boldsymbol{e}$-related terms. The desired EE pose is defined as the initial pose with a predefined z-direction displacement in the EE frame, which specifies the insertion depth.

\section{Experiments}
\subsection{Data Collection and Policy Training}
\subsubsection{Training Data} 
\begin{figure}[h]
	\centering
	\includegraphics[width=3.3in]{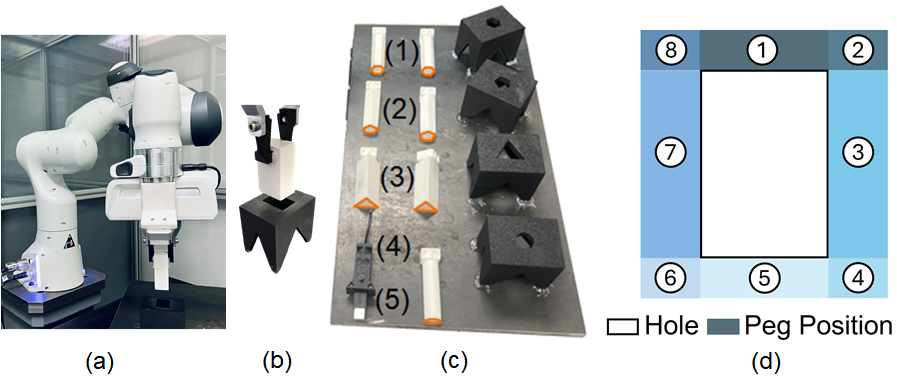}
	\caption{\textbf{(a)} A 7-DoF Franka Emika robot is employed to collect the training demonstrations. The wrench measurements are obtained using the wrist tactile sensor integrated in the Franka robot. \textbf{(b)} All the training data are collected using a 35 mm $\times$ 25 mm $\times$ 60 mm cuboid peg with 0.1 mm clearance.  \textbf{(c)} We evaluate the transferability of the proposed method on five unseen shapes: (1) Hexagonal prism, 60 mm long with a flat-to-flat distance of 19 mm (with 0.1 mm and 0.2 mm clearances); (2) Pentagonal prism, 60 mm long with a flat-to-flat distance of 19 mm (with 0.1 mm and 0.2 mm clearances); (3) Triangular prism, 60 mm long with a side length of 34 mm (with 0.1 mm and 0.2 mm clearances); (4) Cylinder, 60 mm long with a diameter of 20 mm (0.1 mm clearance); and (5) a standard USB-A connector. \textbf{(d)} During the test, the peg is sequentially placed in regions $\textcircled{1} \sim \textcircled{8}$ around the hole. At each region, four misalignment levels (2, 3, 4, and 5 mm) are tested with three repeated trials each, resulting in 12 trials per region.}
     \label{setup1}
\end{figure}

\begin{figure}[t]
	\centering
	\includegraphics[width=3.3in]{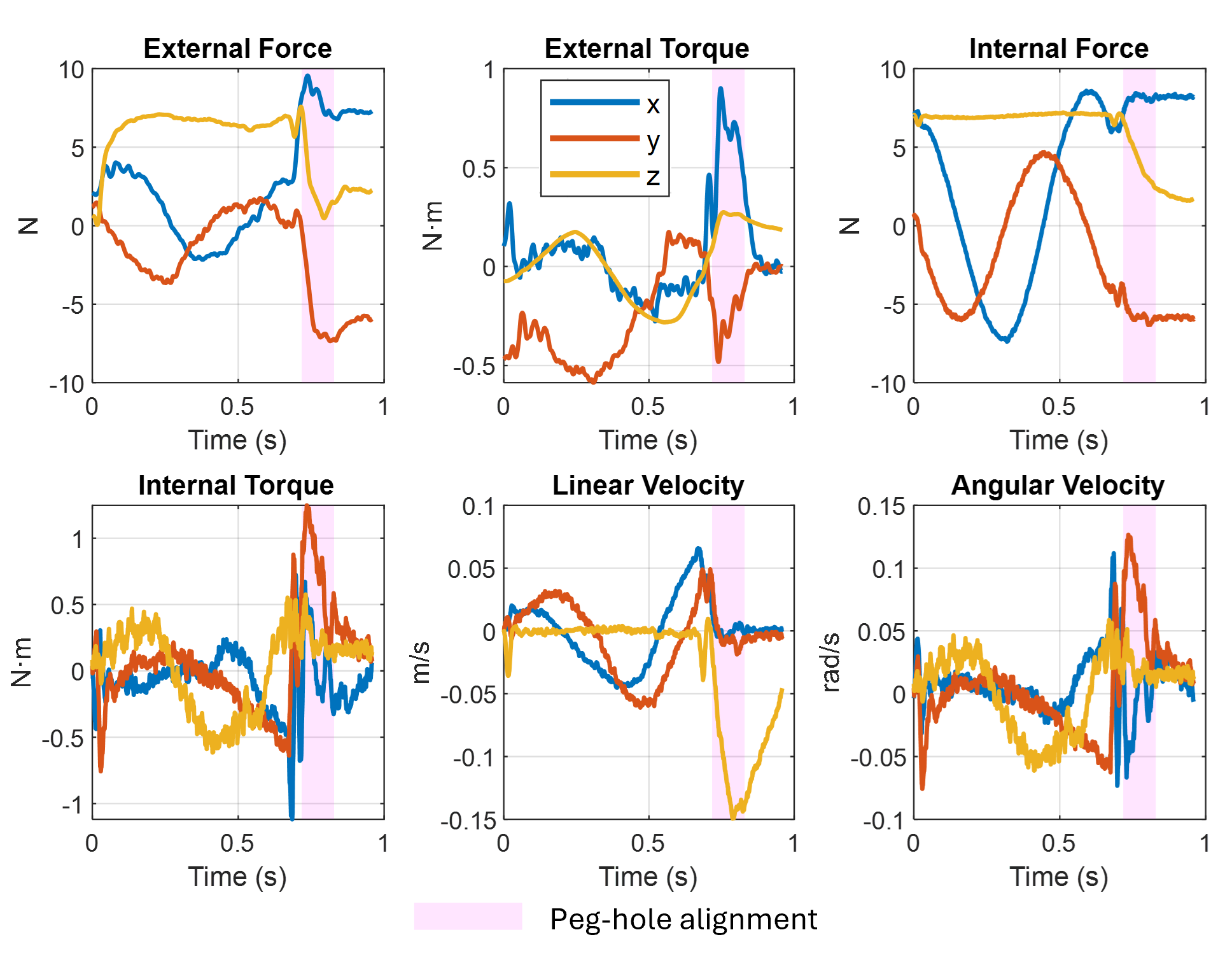}
    
	\caption{An example of the observations in our training data for search. The procedure of peg–hole alignment is highlighted with a light purple background, during which the coaxial force release and the EE’s movement in the z-direction are evident.
    } \label{traindata}

\end{figure}

As shown in Fig. \ref{setup1}(a), we employ a 7-DoF Franka Emika robot to gather the training demonstrations. For the search teacher policy, we rely on \textbf{Algorithm} \ref{algo1} to obtain 300 samples. For each trial, the peg is randomly placed in contact at the rim of the hole within a 5 mm range to cover different initial misalignment conditions. We preserve only the samples that represent successful searches, which, even if not the optimal shortcuts, are achieved with efficient trajectories, while discarding both unsuccessful searches and inefficient successful ones. We record the complete 18-dimensional trajectories of $\boldsymbol{\tau}^{e}_{\mathrm{ext}}$, $\boldsymbol{\tau}^{e}_{\mathrm{in}}$, and $\boldsymbol{v}^{e}$ (refer to Eq. (\ref{extindef})), and concatenate the state of the current frame with that of the previous frame (7 ms earlier) to form the 36-dimensional model input. 
Fig. \ref{traindata} shows an example of observations in our training data for search.
The 6-dimensional trajectory of $\boldsymbol{F}_{ff}$ is also recorded to serve as the ground truth for the model output. All search policy training data are labeled with mode prompt 0. As shown in Fig. \ref{setup1}(b), all the training data come from a single cuboid peg with 0.1 mm clearance, using the same setup as TacDiffusion \cite{tac}.  For the high-precision insertion teacher policy, we utilize the 1,500 training demonstrations provided in TacDiffusion \cite{tac}, all labeled with the mode prompt of 1. They are also acquired using the setup shown in Fig. \ref{setup1}(b). 
\par
\subsubsection{Model Training} We train and evaluate our framework, implemented in PyTorch, with a 48GB NVIDIA RTX A6000 GPU. We trained the model for 2,000 epochs with a batch size of 4,096, using the Adam optimizer with an initial learning rate of $1\times10^{-3}$ scheduled by cosine decay, and set the number of diffusion steps to 50. Considering the imbalance between search and insertion samples (300 \textit{vs}. 1,500), we employ Balanced Batch Sampling (BBS) \cite{bbs} to ensure a balanced representation of each type within mini-batches. 

\begin{figure}[t]
	\centering
	\includegraphics[width=3.3in]{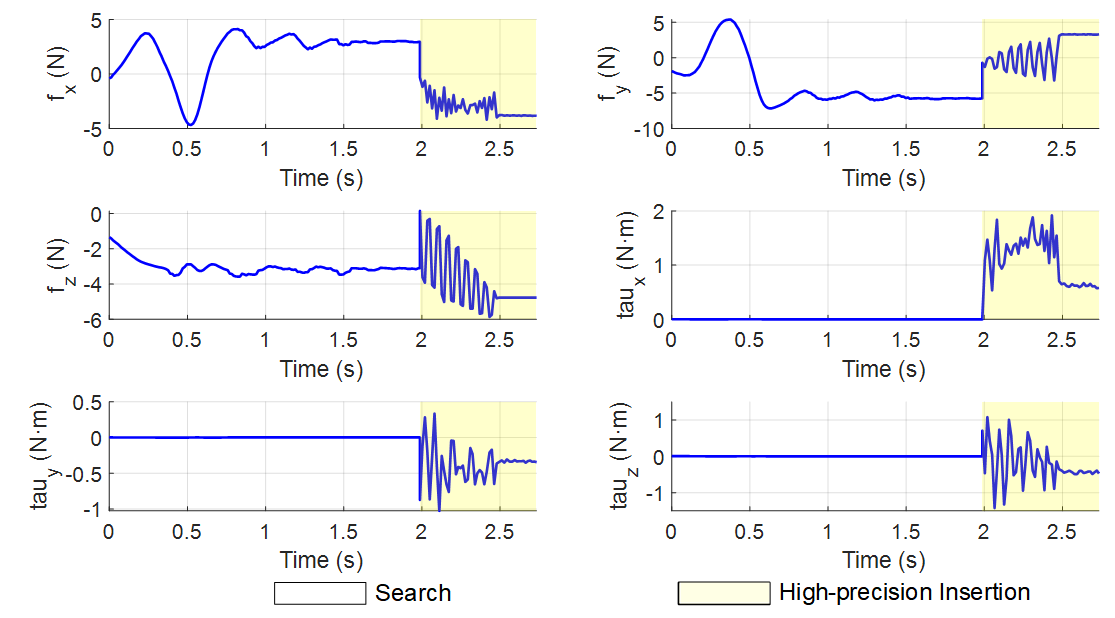}
 
	\caption{An example of the $\boldsymbol{F}_{\mathrm{ff}}$ trajectories generated by our method during a successful peg-in-hole insertion of the pentagonal prism. Note that during the search phase, $\boldsymbol{F}_{\mathrm{ff}}$ shows a varying period and decreasing amplitude, unlike the teacher model.
    }
    
	\label{force}
   
\end{figure}

\begin{table}[t]
\caption{Quantitative comparison with competitors on nine shapes.}
\centering
	\resizebox{1.0\columnwidth}{!}{
  
\begin{tabular}{c|c|c|c|c|c|c|c|c}
\hline 

\multirow{3}{*}{Shape} & \multirow{3}{*}{Method}  & \multicolumn{4}{c|}{Misalignments }  & Suc.& AT.  & AT. \\ \cline{3-6}
  &  & \multicolumn{4}{c|}{(Successes / Total Trials) } &   Rate & All  &  Suc.\\ 
\cline{3-6}
  &  & 2 mm & 3 mm & 4 mm & 5 mm & $(\%)$ & *(s)  &  *(s)\\ 
\hline
\multirow{4}{*}{Cuboid} & TacDiff \cite{tac} & 15/24  & 7/24& 5/24 & 1/24  & 29.1 & 4.53  & 0.96\\ \cline{2-9}
& Teacher-D & 21/24 & 20/24   & 20/24  & 20/24  & 84.3 & 3.17 & 2.65\\ \cline{2-9}
& Teacher-R &22/24  & 22/24  & 21/24  & 20/24  & 88.5 & 3.15 & 2.79\\ \cline{2-9}

& FORGE \cite{forge} & 23/24 & 23/24   & 23/24  & 19/24  & 91.7 & 2.82 & 2.53\\ \cline{2-9}
& \cellcolor{green!30}\textbf{Ours} & 24/24  & 23/24  & 23/24  &22/24  & \cellcolor{green!30}\textbf{95.8}   & \cellcolor{green!30}\textbf{2.55} & 2.40 \\ \hline 
\multirow{2}{*}{Hexagonal} & TacDiff \cite{tac} 
& 15/24 & 7/24 & 3/24 & 0/24 
& 26.0 & 4.55 & 0.92 \\ \cline{2-9}

& Teacher-D 
& 20/24 & 20/24 & 19/24 & 17/24 
& 79.2 & 3.19 & 2.45 \\ \cline{2-9}

\multirow{2}{*}{Prism (0.1)$\dagger$} & Teacher-R 
& 20/24 & 20/24 & 20/24 & 18/24 
& 81.3 & 3.20 & 2.55 \\ \cline{2-9}

& FORGE \cite{forge} 
& 20/24 & 18/24 & 13/24 & 11/24 
& 64.6 & 3.75 & 2.51 \\ \cline{2-9}

& \cellcolor{green!30}\textbf{Ours}  
& 24/24 & 23/24 & 21/24 & 22/24 
& \cellcolor{green!30}\textbf{93.7} 
& \cellcolor{green!30}\textbf{2.65} 
& 2.43 \\ \hline

\multirow{2}{*}{Hexagonal} & TacDiff \cite{tac} & 16/24& 8/24& 3/24 & 0/24& 28.1 & 4.51 & 0.85 \\ \cline{2-9}
& Teacher-D &21/24 & 21/24 & 20/24 & 18/24 & 83.3  & 2.93 & 2.31 \\ \cline{2-9}
\multirow{2}{*}{Prism (0.2)$\dagger$} & Teacher-R &21/24 & 21/24 & 21/24& 19/24 & 85.4 & 2.93 & 2.40\\ \cline{2-9}
& FORGE \cite{forge} & 22/24 & 22/24   & 15/24  & 13/24  & 75.0 & 3.47 & 2.63\\ \cline{2-9}
& \cellcolor{green!30}\textbf{Ours}  & 24/24 & 24/24 & 22/24  & 22/24 & \cellcolor{green!30}\textbf{95.8} & \cellcolor{green!30}\textbf{2.51} & 2.36 \\ \hline

\multirow{2}{*}{Pentagonal} & TacDiff \cite{tac} 
& 16/24 & 7/24 & 2/24 & 1/24 
& 27.1 & 4.62 & 0.91 \\ \cline{2-9}

& Teacher-D 
& 20/24 & 20/24 & 19/24  & 19/24 
& 81.3 & 3.06 & 2.38 \\ \cline{2-9}

\multirow{2}{*}{Prism (0.1)} & Teacher-R 
& 21/24 & 21/24 & 19/24 & 19/24  
& 83.3 & 2.13 & 2.55 \\ \cline{2-9}

& FORGE \cite{forge} 
& 18/24 & 10/24 & 9/24 & 6/24  
& 44.8 & 4.45 & 2.55 \\ \cline{2-9}

& \cellcolor{green!30}\textbf{Ours} 
& 24/24  & 22/24 & 22/24 & 22/24 
& \cellcolor{green!30}\textbf{93.7} 
& \cellcolor{green!30}\textbf{2.39} 
& 2.15 \\ \hline

\multirow{2}{*}{Pentagonal} & TacDiff \cite{tac} & 17/24 & 8/24 & 2/24 & 1/24&  29.1 & 4.49 & 0.83 \\ \cline{2-9}
& Teacher-D & 21/24 & 21/24 & 20/24  & 20/24 & 85.4& 2.77 & 2.27 \\ \cline{2-9}
\multirow{2}{*}{Prism (0.2)} & Teacher-R & 22/24& 22/24 & 20/24 & 20/24  & 87.5 &  2.86 & 2.41\\ \cline{2-9}
& FORGE \cite{forge} & 20/24 & 12/24   & 12/24  & 10/24  & 56.3 & 4.08 & 2.59\\ \cline{2-9}
& \cellcolor{green!30} \textbf{Ours} &24/24  & 23/24 & 22/24 & 22/24 & \cellcolor{green!30}\textbf{94.8} & \cellcolor{green!30}\textbf{2.21} & 2.00\\ \hline

\multirow{2}{*}{Triangular} & TacDiff \cite{tac} 
& 15/24 & 6/24 & 3/24 & 0/24 
& 25.0 & 4.76 & 1.02 \\ \cline{2-9}

& Teacher-D 
& 20/24 & 20/24 & 18/24 & 18/24 
& 79.2 & 3.29 & 2.42 \\ \cline{2-9}

\multirow{2}{*}{Prism (0.1)} & Teacher-R 
& 21/24 & 20/24 & 19/24 & 18/24 
& 81.3 & 3.22 & 2.58 \\ \cline{2-9}

& FORGE \cite{forge} & 13/24 & 11/24   & 9/24  & 7/24  & 41.7 & 4.62 & 2.71\\ \cline{2-9}

& \cellcolor{green!30} \textbf{Ours} 
& 23/24 & 23/24 & 21/24 & 20/24 
& \cellcolor{green!30}\textbf{90.6} 
& \cellcolor{green!30}\textbf{2.56} 
& 2.20 \\ \hline

\multirow{2}{*}{Triangular} & TacDiff \cite{tac} & 17/24 & 7/24 & 3/24& 0/24& 28.1 & 4.57 & 0.93 \\ \cline{2-9}

& Teacher-D & 21/24& 21/24 & 19/24 & 19/24 & 83.3 & 2.91 & 2.30 \\ \cline{2-9}
\multirow{2}{*}{Prism (0.2)} & Teacher-R &22/24 & 21/24 & 20/24 & 19/24 & 85.4 & 2.96 & 2.44\\ \cline{2-9}
& FORGE \cite{forge} & 13/24 & 11/24   & 12/24  & 8/24  & 45.8 & 4.48 & 2.69\\ \cline{2-9}
& \cellcolor{green!30}\textbf{Ours} & 23/24 & 22/24 & 21/24 & 21/24 & \cellcolor{green!30}\textbf{90.6} & \cellcolor{green!30}\textbf{2.43} & 2.06\\ \hline

\multirow{4}{*}{Cylinder} & TacDiff \cite{tac} & 16/24& 6/24 & 4/24 & 2/24& 29.2 & 4.48 & 0.80\\ \cline{2-9}

& Teacher-D & 22/24  & 20/24 & 20/24 & 19/24  & 84.4  &  2.59 & 1.95 \\ \cline{2-9}
& Teacher-R &22/24 & 21/24 & 21/24 & 21/24 & 88.5 & 2.58 & 2.13\\ \cline{2-9}
& FORGE \cite{forge} & 22/24 & 23/24   & 20/24  & 20/24  & 88.5 & 2.55 & 2.10\\ \cline{2-9}
& \cellcolor{green!30} \textbf{Ours} & 24/24 & 24/24 & 23/24 & 23/24 & \cellcolor{green!30}\textbf{97.9}  & \cellcolor{green!30}\textbf{1.85} & 1.76 \\ \hline

\multirow{4}{*}{USB-A} & TacDiff \cite{tac} & 17/24 & 9/24 & 5/24& 2/24& 34.3 & 4.38 & 1.30\\ \cline{2-9}

& Teacher-D & 21/24 & 20/24 & 19/24& 19/24 & 82.3 &  3.40 & 2.84 \\ \cline{2-9}
& Teacher-R &22/24 & 20/24 & 20/24 & 18/24 & 83.3 & 3.39 &2.94\\ \cline{2-9}
& FORGE \cite{forge} & 22/24 & 21/24   & 21/24  & 20/24  & 87.5 & 3.17 & 2.77\\ \cline{2-9}
& \cellcolor{green!30}\textbf{Ours} & 22/24 &22/24 &22/24 &21/24 &  \cellcolor{green!30}\textbf{90.6}  & \cellcolor{green!30}\textbf{2.94} & 2.74 \\ \hline
\multicolumn{9}{l}{* AT. All: average time (including failures); AT. Suc.: average time (success-only).} \\
\multicolumn{9}{l}{$\dagger$ (0.1) and (0.2) denote 0.1 mm and 0.2 mm clearance, respectively.} 
\end{tabular}
}
\label{tab1}

\end{table}

\subsection{Experimental Setup}

\subsubsection{Competitors} We compare the proposed SI-Diff (\textbf{Ours}) against TacDiffusion \cite{tac}, Teacher-D, Teacher-R, and FORGE \cite{forge}. 
Teacher-D and Teacher-R are two variants constructed based on \textbf{Algorithm 1}. In Teacher-D, the parameters of Algorithm 1 are fixed \footnote{$p_s=1$, $d_x=1$, $d_y=1$, $a= -0.15$, $b= -0.15$, $c=7.0$, $\theta=-50.0$, and $\Delta \theta = -5.5e^{-3}.$}, whereas in Teacher-R, the parameters are randomly sampled.
Both variants combine Algorithm 1 with vanilla TacDiffusion, where the former handles the search stage, and the latter performs high-precision insertion. As such, they can be regarded as TacDiffusion augmented with the proposed rule-based search module, serving as upgraded versions of 1kHz-BT \cite{tree}.
FORGE \footnote{ We use the official FORGE code implementation together with our cuboid peg and hole models for training. Following its paper, we extend the noisy estimate of the fixed part’s pose from 3D translation to the full 6D pose in the code implementation. We also revised the keypoint definition from colinear points to coplanar corner points with associated normal directions. Finally, we apply a curriculum learning strategy (cuboid peg; clearance: $1\,\mathrm{mm} \rightarrow 0.4\,\mathrm{mm} \rightarrow 0.1\,\mathrm{mm}$). Otherwise, the model cannot be trained successfully.}is a state-of-the-art force-guided reinforcement learning–based method.

\begin{figure}[t]
	\centering
	\includegraphics[width=3.3in]{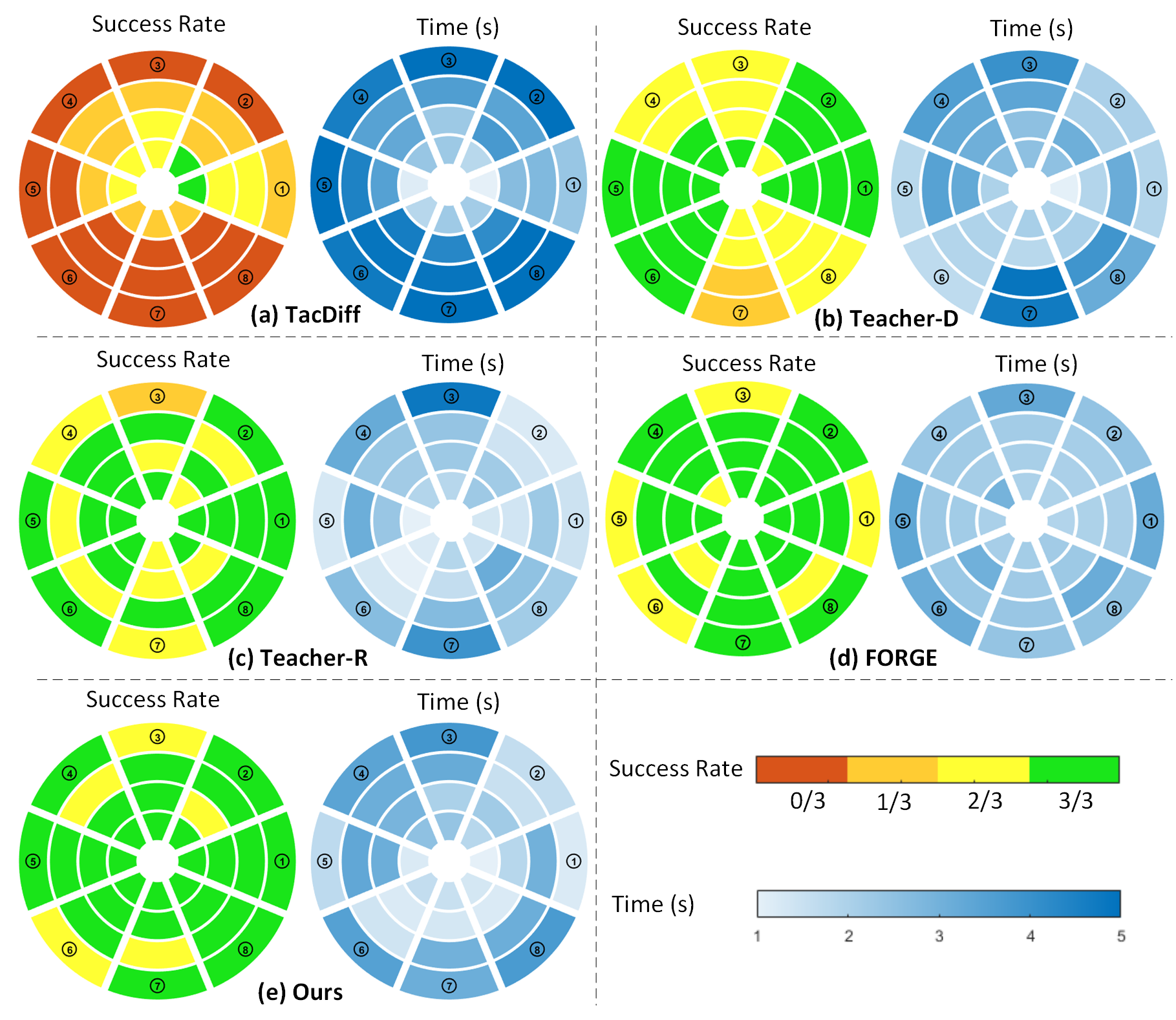}
	\caption{Heatmaps illustrating the distributions of success rates and task efficiency in tests with the cuboid peg: (a) TacDiffusion \cite{tac}, (b) Teacher-D, (c) Teacher-R, (d) FORGE \cite{forge}, and (e) Ours. 
    }
	\label{hotmap}
\end{figure}
\subsubsection{Experimental Procedure} We first follow \cite{tac} to construct the feedforward-force-based impedance controller (see Eq. (\ref{controller})), where
$\mathbf{K}$ and $\mathbf{D}$ are defined as diagonal matrices with diagonal entries of 523.91 and 24.98, respectively. 
Note that the same controller is applied to the four competitors for a fair comparison. Namely, the only difference among the competitors is the manner in which $\boldsymbol{F}_{ff}$ is generated.
For TacDiffusion \footnote{We apply the officially released trained weights of TacDiffusion, while $\mathbf{K}$ and $\mathbf{D}$ are the only hyperparameters that need to be specified.}, which consists of a single stage, the desired EE pose is defined as the initial pose with a 2 cm displacement along the z-direction of the EE frame (11.5 mm for the USB-A). 
Teacher-D, Teacher-R, and Ours each consist of two stages and rely on the displacement along the z-direction of the EE frame as the signal for mode switching. {As described in Algorithm \ref{algo1}, once the displacement reaches 3 mm, the search stage is completed and the mode switches to insertion. The insertion stage continues for an additional 1.7 cm \footnote{Note that in the search stage, the desired pose (which determines 
$\boldsymbol{e}$) corresponds to the pose at which the search begins, whereas in the insertion stage, the desired pose is shifted by -1.7 cm along the z-axis relative to the pose at which insertion begins. The robot’s motion is regulated by Eqs. (\ref{dynamic}) and (\ref{controller}) throughout the entire process.} corresponding to a total displacement of 2 cm (8.5 mm for the USB-A).}
Moreover, since the inference frequency of our model (101.62 Hz) and TacDiffusion (123.90 Hz) is lower than the 1 kHz control rate required by the Franka robot, we follow \cite{tac} and apply a second-order low-pass filter to generate 1 kHz control commands.
As shown in Fig. \ref{setup1}(c), in addition to the cuboid peg used for collecting training demonstrations, we also evaluate the transferability of our method on five unseen shapes: hexagonal prism, pentagonal prism, triangular prism, cylinder, and USB-A. As shown in Fig. \ref{setup1}(d), the peg is sequentially placed in eight regions with initial misalignment conditions of 2, 3, 4, and 5 mm. The misalignment is measured as the translational offset between the peg and hole centers in the x-y plane. 96 trials are conducted for each shape and method. A 6 s cut-off is applied: if the z-direction displacement remains unchanged for 6 s, the trial is deemed a failure and its execution time is recorded as 6 s. 
\subsubsection{Metric} We use the success rate (successful cases over total trials) to compare methods. To assess efficiency, we report average execution times both overall (failure-included) and success-only. Since failure trials are recorded as 6 s, a method with higher success rates could appear faster in the failure-included average. Reporting the success-only average mitigates this bias.

\subsection{Experimental Results and Analysis}
\subsubsection{Qualitative Results}
We present the qualitative experimental results in Fig. \ref{force}, which shows an example of the $\boldsymbol{F}_{ff}$ trajectories generated by our method during a successful peg-in-hole insertion of the pentagonal prism.
As seen, in the search stage, the model outputs oscillatory $f_x$ and $f_y$ resembling sinusoidal patterns to regulate search motion, while keeping $f_z$ constant and producing no torque. When the search is done, the model automatically maintains the last $\boldsymbol{F}_{ff}$. It should be noted that although the search teacher policy can only demonstrate sinusoidal $\boldsymbol{F}_{ff}$ with a strictly increasing amplitude, a fixed frequency, and a fixed clockwise search direction (see the equations of $f_x$ and $f_y$ in Algorithm \ref{algo1}), our model exhibits action behaviors with varying amplitude, frequency, and search direction, and thereby demonstrating more flexible and adaptive search strategies. It is evident that the amplitudes decrease toward the end of the search, as the peg and hole gradually become aligned and the required EE movements become progressively smaller. Two mechanisms underlie this behavior. First, the demonstrations provide paired tactile observations and actions, enabling the model to recognize which tactile patterns indicate that the search is approaching completion (see Fig. \ref{traindata}). Second, the large mini-batch size (4096) used during training allows the model to access multiple demonstrations simultaneously, enabling it to learn variations in amplitude, frequency, and search direction within each mini-batch. In the high-precision insertion stage, the model generates sinusoidal $f_x$, $f_y$, $\tau_x$, $\tau_y$, and $\tau_z$ to escape from stuck conditions while maintaining a consistent $f_z$. When insertion is completed, the last $\boldsymbol{F}_{ff}$ is automatically preserved.
These distinct behaviors are produced through a single framework with the same weights, thanks to the proposed force-domain diffusion policy with mode conditioning.

\subsubsection{Quantitative Results}
The quantitative comparison with competitors on six shapes is shown in Table \ref{tab1} and Fig. \ref{hotmap}. As seen from the results in Table \ref{tab1}, TacDiffusion \cite{tac} cannot handle initial misalignments exceeding 2 mm, and the overall success rate for each shape remains below 35\%. It achieves the fastest success-only average execution time because its strategy relies on high-frequency wiggling, a motion pattern that can sometimes, by chance, quickly drive the peg into the hole. However, the success-only average execution time carries little significance when the success rate is low.
In contrast, both Teacher-D and Teacher-R reach success rates of around 83\%$\sim$88\%, illustrating the necessity of the search motion. 
Teacher-R shows a slightly better success rate and failure-included average execution time than Teacher-D, while the latter achieves a slightly better success-only average execution time.
This is because Teacher-D uses fixed parameters, resulting in a moderate and consistent search strategy. In contrast, Teacher-R produces greater trajectory diversity due to parameter randomization in Algorithm \ref{algo1}. Some of these trajectories correspond to conservative search strategies that yield a higher probability of success but at the cost of reduced efficiency.
FORGE performs well on the training shape and also transfers well to unseen cylinder peg and USB-A. However, its performance declines on other unseen shapes. This suggests that the learned shape-specific priors can generalize to a geometrically simple shape (cylinder) or a large-clearance task (e.g., USB-A), but transfer less effectively to a more complex geometry with small clearance.
Our method yields the best performance on all objects, including the highest success rate and the best efficiency. In particular, by comparing the success-only average execution time, we can confirm that the best overall average execution time is not merely an artifact of the higher success rate. This highlights the benefit of learning from the successful and efficient demonstrations of Teacher-R. Moreover, the proposed method exhibits good transferability to unseen objects.
In Fig. \ref{hotmap}, we present heatmaps illustrating the distributions of success rates and efficiency for the methods. As seen, TacDiffusion and Teacher-D have evident distributions of lower success rates and efficiency, whereas Teacher-R, FORGE, and Ours exhibit relatively more random distributions. The performance of TacDiffusion may be influenced by potential biases in the training data. The biased distribution of Teacher-D is illustrated in Fig. \ref{vanilla}.
\begin{table}[t]
\caption{ The ablation study of the proposed framework.}
\centering
	\resizebox{1.0\columnwidth}{!}{
\begin{tabular}{ccc|c|c}
\hline
 Randomness in &Mode  &Balanced   & Overall  & Average  \\
 Search Teacher Policy&Embedding & Batch Sampling & ($\%$) & Time (s) \\
\hline 
&\ding{51} &  \ding{51} & 82.3 & 3.22 \\ 
\ding{51} && \ding{51} & 40.6 & 4.23 \\

\ding{51} & \ding{51} &  & 92.7 & 2.61 \\ 
\ding{51} & \ding{51} & \ding{51}  & 95.8 & 2.55 \\ \hline
\end{tabular}
}
\label{tab2}
\end{table}

\subsection{Ablation Study}
To verify the effectiveness of our key components, we conduct ablation studies on the cuboid peg, as shown in Table \ref{tab2}. The first row corresponds to SI-Diff trained with Teacher-D, where mode embedding and BBS are retained but randomness in the search teacher policy is removed. Compared with Teacher-D in Table \ref{tab1}, the performance is even worse, since Teacher-D follows a fixed search trajectory independent of external or internal wrenches, which decouples tactile observations from actions and weakens learning.
The second row shows training with the same data and BBS but without mode embedding. Although equal numbers of search and insertion demonstrations are used, the model still struggles to learn the two distinct behaviors. Performance improves over TacDiffusion but remains inferior to ours.
Comparing the third and fourth rows demonstrates the effectiveness of BBS under imbalanced search and insertion data.

\subsection{Limitations and Discussions}
Our approach has two limitations. First, mode switching between search and insertion is triggered by z-displacement rather than being learned from tactile cues. Second, we focus only on x–y misalignment and small-clearance insertion, and extending the model to correct out-of-plane and yaw errors would further improve robustness.
Although TacDiffusion \cite{tac} shows low performance in Table \ref{tab1} under large misalignment, it achieves over 90\% success when the initial error is within 1 mm. Moreover, its wiggling strategy remains essential for high-precision insertion, as applying $f_z$ alone is insufficient for small clearances \cite{tac,tree}.

\section{Conclusions}
In this work, we develop a framework for learning search and high-precision insertion through a force-domain diffusion policy, dubbed SI-Diff. We first design a new mode-conditioning mechanism that enables the diffusion model to simultaneously learn and perform the distinct action behaviors of search and high-precision insertion. We then propose a search teacher policy that can generate diverse search trajectories. By learning from the successful and efficient demonstrations of the teacher policy, our model surpasses both the teacher policy and the state-of-the-art baseline, TacDiffusion \cite{tac}, in terms of success rate and efficiency. We conduct thorough experiments to demonstrate the effectiveness of the proposed framework and the good zero-shot transferability of our method to unseen shapes.
Compared to TacDiffusion, which tolerates only 2 mm misalignment, our method extends the tolerance to 5 mm, enhancing robustness in real-world peg-in-hole tasks.
	%

	\bibliographystyle{IEEEtran} 
	\bibliography{reference}

\end{document}